\documentclass[11pt,a4paper]{article}
\usepackage[hyperref]{acl2021}
\usepackage{times}
\usepackage{latexsym}

\usepackage{microtype}
\usepackage{amsmath}
\usepackage{amsfonts}
\usepackage{microtype}
\usepackage{graphicx}
\usepackage{multirow}
\usepackage{booktabs}
\usepackage{hhline}
\usepackage{xspace}
\usepackage{color}
\usepackage{enumitem}
\setlist[itemize]{leftmargin=*}
\setitemize[1]{itemsep=0pt,partopsep=0pt,parsep=\parskip,topsep=0pt}
\DeclareSymbolFont{extraup}{U}{zavm}{m}{n}
\DeclareMathSymbol{\varheart}{\mathalpha}{extraup}{86}
\DeclareMathSymbol{\vardiamond}{\mathalpha}{extraup}{87}
\usepackage{amssymb}
\usepackage{pifont}
\aclfinalcopy 

\title{Cross-modal Memory Networks for Radiology Report Generation}

\author{
    Zhihong Chen$^{\spadesuit\heartsuit}$, \hspace{0.2cm}
    Yaling Shen$^{\spadesuit}$, \hspace{0.2cm}
    Yan Song$^{{\spadesuit}\heartsuit\dag}$, \hspace{0.2cm}
    Xiang Wan$^{\heartsuit}$ \\
    $^{\spadesuit}$The Chinese University of Hong Kong (Shenzhen)\\
    $^{\heartsuit}$Shenzhen Research Institute of Big Data\\
    $^{\spadesuit}$\texttt{\{zhihongchen,yalingshen\}@link.cuhk.edu.cn}\\
    $^{\spadesuit}$\texttt{songyan@cuhk.edu.cn} \hspace{0.2cm}
    $^{\heartsuit}$\texttt{wanxiang@sribd.cn}
}

\date{}

\begin{document}
\maketitle
\renewcommand{\thefootnote}{\fnsymbol{footnote}}
\footnotetext[2]{Corresponding author.}
\renewcommand{\thefootnote}{\arabic{footnote}}

\begin{abstract}
Medical imaging plays a significant role in clinical practice of medical diagnosis,
where the text reports of the images are essential in understanding them and facilitating later treatments.
By generating the reports automatically, it is beneficial to help lighten the burden of radiologists and significantly promote clinical automation, which already attracts much attention in applying artificial intelligence to medical domain.
Previous studies mainly follow the encoder-decoder paradigm and focus on the aspect of text generation, with few studies considering the importance of cross-modal mappings and explicitly exploit such mappings to facilitate radiology report generation.
In this paper, we propose a cross-modal memory networks (CMN) to enhance the encoder-decoder framework for radiology report generation, where a shared memory is designed to record the alignment between images and texts so as to facilitate the interaction and generation across modalities.
Experimental results illustrate the effectiveness of our proposed model, where state-of-the-art performance is achieved on two widely used benchmark datasets, i.e., IU X-Ray and MIMIC-CXR.
Further analyses also prove that our model is able to better align information from radiology images and texts so as to help generating more accurate reports in terms of clinical indicators.\footnote{Our code and the best performing models are released at \url{https://github.com/zhjohnchan/R2GenCMN}.}
\end{abstract}

\section{Introduction}
Interpreting radiology images (e.g., chest X-ray) and writing diagnostic reports are essential operations in clinical practice and normally requires considerable manual workload.
Therefore, radiology report generation, which aims to automatically generate a free-text description based on a radiograph, is highly desired to ease the burden of radiologists while maintaining the quality of health care.
Recently, substantial progress has been made towards research on automated radiology report generation models \cite{coatt,hrgr,mimic,clinically,cmas}.
Most existing studies adopt a conventional encoder-decoder architecture, with convolutional neural networks (CNNs) as the encoder and recurrent (e.g., LSTM/GRU) or non-recurrent networks (e.g., Transformer) as the decoder following the image captioning paradigm \cite{showandtell,updown}.
\begin{figure}[t]
\centering
\includegraphics[width=0.47\textwidth, trim=0 20 0 20]{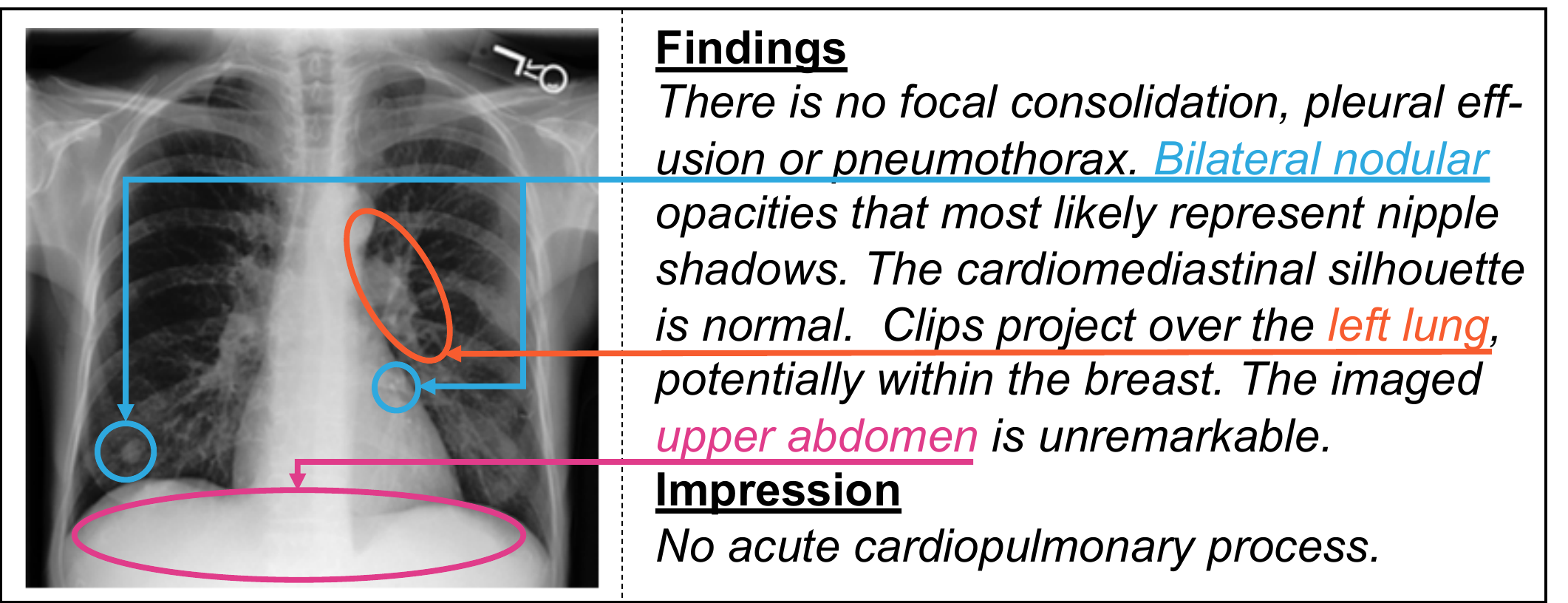}
\caption{A chest X-ray image and its report including findings and impression, where aligned visual and textual features are marked in different colors.}
\label{fig:example}
\vskip -1.7em
\end{figure}
\begin{figure*}[t]
\centering
\includegraphics[width=0.98\textwidth, trim=0 10 0 0]{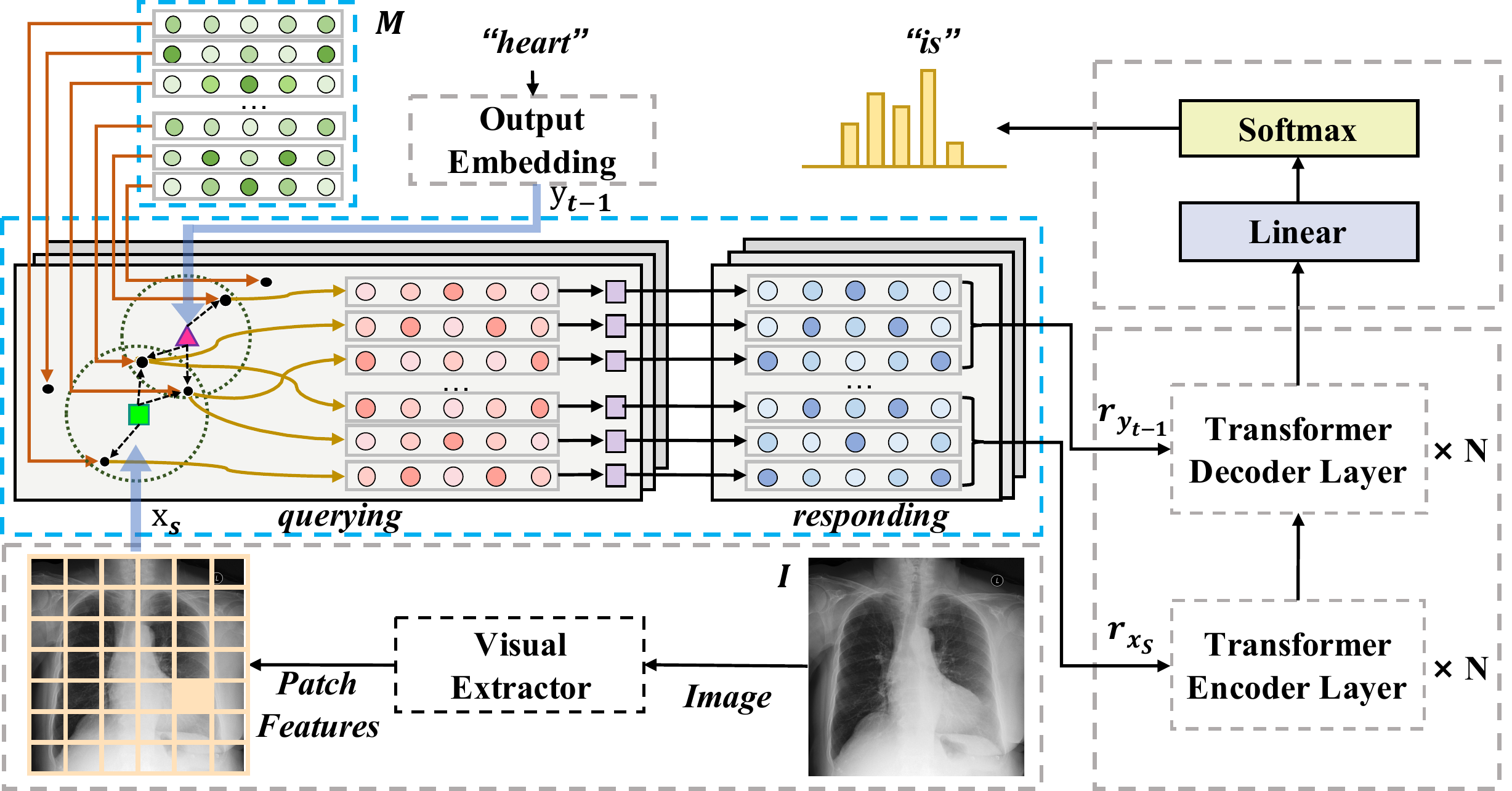}
\caption{The overall architecture of our proposed approach, where the visual extractor, encoder and decoder are shown in gray dash boxes with the details omitted.
The cross-modal memory networks are illustrated in blue dash boxes with presenting the detailed process of memory querying and responding.
}
\label{fig:model-architecture}
\vskip -1em
\end{figure*}
Although these methods have achieved remarkable performance, they are still restrained in fully employing the information across radiology images and reports, such as the mappings demonstrated in Figure \ref{fig:example} that aligned visual and textual features point to the same content.
The reason for the restraint comes from both the limitation of annotated correspondences between image and text for supervised learning as well as the lack of good model design to learn the correspondences.
Unfortunately, few studies\footnote{Along this research track, recently there is only \citet{coatt} studying on 
a multi-task learning framework with a co-attention mechanism to explicitly explore information linking particular parts in a radiograph and its corresponding report.} are dedicated to solving the restraint.
Therefore, it is expected to have a better solution to model the alignments across modalities and further improve the generation ability, although promising results are continuously acquired by other approaches \cite{hrgr,clinically,cmas,r2gen}.

In this paper, we propose an effective yet simple approach to radiology report generation enhanced by cross-modal memory networks (CMN), which is designed to facilitate the interactions across modalities (i.e., images and texts).
In detail, we use a memory matrix to store the cross-modal information and use it to perform memory querying and memory responding for the visual and textual features, where for memory querying, we extract the most related memory vectors from the matrix and compute their weights according to the input visual and textual features, and then generate responses by weighting the queried memory vectors.
Afterwards, the responses corresponding to the input visual and textual features are fed into the encoder and decoder, so as to generate reports enhanced by such explicitly learned cross-modal information.
Experimental results on two benchmark datasets, \textsc{IU X-Ray} and \textsc{MIMIC-CXR}, confirm the validity and effectiveness of our proposed approach, where state-of-the-art performance is achieved on both datasets.
Several analyses are also performed to analyze the effects of different factors affecting our model, showing that our model is able to generate reports with meaningful image-text mapping while requiring few extra parameters in doing so.

\section{The Proposed Approach}
We regard radiology report generation as an image-to-text generation task, for which there exist several solutions \cite{showandtell,xu2015show,updown,m2}.
Although images are organized as 2-D format,
we follow the standard sequence-to-sequence paradigm for this task as that performed in \citet{r2gen}.
In detail, the source sequence is $\mathbf{X}=\{\mathbf{x}_1,\mathbf{x}_2,...,\mathbf{x}_s,...,\mathbf{x}_S\}$, where $\mathbf{x}_s\in \mathbb{R}^d$ are extracted by visual extractors from a radiology image $\mathbf{I}$ and the target sequence are the corresponding report $\mathbf{Y}=\{y_1, y_2,...,y_t,...,y_T\}$, where $y_t\in\mathbb{V}$ are the generated tokens, $T$ the length of the report and $\mathbb{V}$ the vocabulary of all possible tokens.
The entire generation process is thus formalized as a recursive application of the chain rule
\begin{equation}
\setlength\abovedisplayskip{4pt}
\setlength\belowdisplayskip{4pt}
    p(\mathbf{Y}|\mathbf{I}) = \prod_{t=1}^{T} p(y_{t}|y_{1}, ..., y_{t-1}, \mathbf{I})
\end{equation}
The model is then trained to maximize $p(\mathbf{Y}|\mathbf{I})$ through the negative conditional log-likelihood of $\mathbf{Y}$ given the $\mathbf{I}$:
\begin{equation}
\setlength\abovedisplayskip{4pt}
\setlength\belowdisplayskip{4pt}
    \theta^{*} = \mathop{\arg\max}_{\theta} \sum_{t=1}^{T} \log p(y_{t}|y_{1}, ..., y_{t-1}, \mathbf{I};\theta)
\end{equation}
where $\theta$ is the parameters of the model.
An overview of the proposed model is demonstrated in Figure \ref{fig:model-architecture}, with cross-modal memories emphasized.
The details of our approach are described in following subsections regarding to its
three major components, i.e., the visual extractor, the cross-modal memory networks and the encoder-decoder process enhanced by the memory.

\subsection{Visual Extractor}
To generate radiology reports, the first step is to extract the visual features from radiology images.
In our approach, the visual features $\mathbf{X}$ of a radiology image $\mathbf{I}$ are extracted by pre-trained convolutional neural networks (CNN), such as VGG \cite{vgg} or ResNet \cite{resnet}.
Normally, an image is decomposed into regions of equal size\footnote{E.g.,
VGG/ResNet uses region size 32 $\times$ 32 (in pixels).},
i.e., patches, and the features (representations) of them are extracted from the last convolutional layer of CNN.
Once extracted, the features in our study are expanded into a sequence by concatenating them from each row of the patches on the image.
The resulted representation sequence is used as the source input for all subsequent modules and the process is formulated as
\begin{equation}
\setlength\abovedisplayskip{6pt}
\setlength\belowdisplayskip{6pt}
   \{\mathbf{x}_1,\mathbf{x}_2,...,\mathbf{x}_s,...,\mathbf{x}_S\}=f_v(\mathbf{I})
\end{equation}
where $f_v(\cdot)$ refers to the visual extractor.

\subsection{Cross-modal Memory Networks}
To model the alignment between image and text, existing studies tend to map between images and texts directly from their encoded representations (e.g., \citet{coatt} used a co-attention to do so).
However, this process always suffers from the limitation that the representations across modalities are hard to be aligned, so that an intermediate medium is expected to enhance and smooth such mapping.
To address the limitation,
we propose to use CMN to better model the image-text alignment,
so as to facilitate the report generation process.

With using the proposed CMN, the mapping and encoding can be described in the following procedure.
Given a source sequence $\{\mathbf{x}_1,\mathbf{x}_2,...,\mathbf{x}_S\}$ (features extracted from the visual extractor) from an image, we feed it to this module to obtain the memory responses of the visual features $\{\mathbf{r}_{\mathbf{x}_1}, \mathbf{r}_{\mathbf{x}_2},...,\mathbf{r}_{\mathbf{x}_S}\}$.
Similarly, given a generated sequence $\{y_1, y_2,..., y_{t-1}\}$ with its embedding $\{\mathbf{y}_1,\mathbf{y}_2,...,\mathbf{y}_{t-1}\}$, it is also fed to the cross-modal memory networks to output the memory responses of the textual features $\{\mathbf{r}_{\mathbf{y}_1}, \mathbf{r}_{\mathbf{y}_2},...,\mathbf{r}_{\mathbf{y}_{t-1}}\}$.
In doing so, the shared information of visual and textual features can be recorded in the memory so that the entire learning process is able to explicitly map between the images and texts.
Specifically, the cross-modal memory networks employs a matrix to preserve information for encoding and decoding process, where each row of the matrix (i.e., a memory vector) records particular cross-modal information connecting images and texts.
We denote the matrix as $\mathbf{M} = \{\mathbf{m}_1,\mathbf{m}_2,...,\mathbf{m}_i,...,\mathbf{m}_{\mathcal{N}}\}$, where $\mathcal{N}$ represents the number of memory vectors and $\mathbf{m}_i \in \mathbb{R}^{d}$ the memory vector at row $i$ with $d$ referring to its dimension.
During the process of report generation, CMN is operated with two main steps, namely, querying and responding, whose details are described as follows.\footnote{Note that these two steps are performed in both training and inference stages, where in inference, all textual features are obtained along with the generation process.}

\vskip 0.5em
\noindent\textbf{Memory Querying}~
We apply multi-thread\footnote{Thread number can be arbitrarily set in experiments.} querying to perform this operation,
where in each thread the querying process follows the same procedure described as follows.

In querying memory vectors, the first step is to ensure the input visual and textual features are in the same representation space.
Therefore, 
we convert each memory vector in $\mathbf{M}$ as well as input features through linear transformation by
\begin{align}
\setlength\abovedisplayskip{6pt}
\setlength\belowdisplayskip{6pt}
    \mathbf{k}_i&=\mathbf{m}_i\cdot\mathbf{W}_{k}\\
    \mathbf{q}_{s}&=\mathbf{x}_{s}\cdot\mathbf{W}_{q}\\
    \mathbf{q}_{t}&=\mathbf{y}_{t}\cdot\mathbf{W}_{q}
\end{align}
where $\mathbf{W}_{k}$ and $\mathbf{W}_{q}$ are trainable weights for the conversion.
Then we separately extract the most related memory vector to visual and textual features according to their distances $D_{s_{i}}$ and $D_{t_{i}}$ through
\begin{align}
\setlength\abovedisplayskip{6pt}
\setlength\belowdisplayskip{6pt}
    D_{s_{i}} &= \frac{\mathbf{q}_{s} \cdot \mathbf{k}_{i}^{\top}}{\sqrt{d}}\\
    D_{t_{i}} &= \frac{\mathbf{q}_{t} \cdot \mathbf{k}_{i}^{\top}}{\sqrt{d}}
\end{align}
where the number of extracted memory vectors can be controlled by a hyper-parameter $\mathcal{K}$ to regularize how much memory is used.
We denote the queried memory vectors as $\{\mathbf{k}_{s_{1}}, \mathbf{k}_{s_{2}}, ..., \mathbf{k}_{s_{j}}, ..., \mathbf{k}_{s_{\mathcal{K}}\}}$ and $\{\mathbf{k}_{t_{1}}, \mathbf{k}_{t_{2}}, ..., \mathbf{k}_{t_{j}}, ..., \mathbf{k}_{t_{\mathcal{K}}\}}$.
Afterwards,
the importance weight of each memory vector with respect to visual and textual features are obtained by normalization over all distances by
\begin{align}
\setlength\abovedisplayskip{6pt}
\setlength\belowdisplayskip{6pt}
    w_{s_{i}} &= \frac{\text{exp}(D_{s_{i}})}{\Sigma_{j=1}^{\mathcal{K}} \text{exp}(D_{s_{j}})}\\
    w_{t_{i}} &= \frac{\text{exp}(D_{t_{i}})}{\Sigma_{j=1}^{\mathcal{K}} \text{exp}(D_{t_{j}})}
\end{align}
Note that the above steps are applied in each thread
to allow memory querying from different memory representation subspaces.

\vskip 0.5em
\noindent\textbf{Memory Responding}~
The responding process is also conducted in a multi-thread manner corresponding to the query process.
For each thread,
we firstly perform a linear transformation on the queried memory vector via
\begin{equation}
    \mathbf{v}_{i}=\mathbf{m}_{i}\cdot\mathbf{W}_{v}
\end{equation}
where $\mathbf{W}_{v}$ is the trainable weight for $\mathbf{m}_{i}$.
So that all memory vectors $\{\mathbf{v}_{s_{1}}, \mathbf{v}_{s_{2}}, ..., \mathbf{v}_{s_{j}}, ..., \mathbf{v}_{s_{\mathcal{K}}\}}$ are transferred into 
$\{\mathbf{v}_{t_{1}}, \mathbf{v}_{t_{2}}, ..., \mathbf{v}_{t_{j}}, ..., \mathbf{v}_{t_{\mathcal{K}}\}}$.
Then, we obtain the memory responses for visual and textual features by weighting over the transferred memory vectors by
\begin{align}
    \mathbf{r}_{\mathbf{x}_s} &= \Sigma_{i=1}^{\mathcal{K}} w_{s_{i}} \mathbf{v}_{s_{i}} \\
    \mathbf{r}_{\mathbf{y}_t} &= \Sigma_{i=1}^{\mathcal{K}} w_{t_{i}} \mathbf{v}_{t_{i}}
\end{align}
where $w_{s_{i}}$ and $w_{t_{i}}$ are the weights obtained from memory querying.
Similar to memory querying, we apply memory responding to all the threads so as to obtain responses from different memory representation subspaces.

\subsection{Encoder-Decoder}

Since the quality of input representation plays an important role in model performance \cite{pennington-etal-2014-glove,song-etal-2017-learning,song-etal-2018-directional,peters-etal-2018-deep,ijcai2018-607,devlin2019bert,song2021zen},
the encoder-decoder in our model is built upon standard Transformer (which is a powerful architecture that achieved state-of-the-art in many tasks), where memory responses of visual and textual features are functionalized as the input of the encoder and decoder so as to enhance the generation process.
In detail, as the first step, the memory responses $\{\mathbf{r}_{\mathbf{x}_1}, \mathbf{r}_{\mathbf{x}_2},...,\mathbf{r}_{\mathbf{x}_S}\}$ for visual features are fed into the encoder through
\begin{equation}
    \{\mathbf{z}_1,\mathbf{z}_2,...,\mathbf{z}_S\}=f_e(\mathbf{r}_{\mathbf{x}_1},\mathbf{r}_{\mathbf{x}_2},...,\mathbf{r}_{\mathbf{x}_S})
\end{equation}
where $f_e(\cdot)$ represents the encoder.
Then the resulted intermediate states $\{\mathbf{z}_1,\mathbf{z}_2,...,\mathbf{z}_S\}$ are sent to the decoder at each decoding step, jointly with the 
memory responses $\{\mathbf{r}_{\mathbf{y}_1}, \mathbf{r}_{\mathbf{y}_2},...,\mathbf{r}_{\mathbf{y}_{t-1}}\}$ for the textual features of generated tokens from previous steps, so as to generate the current output $y_t$ by
\begin{equation}
    y_{t}=f_d(\mathbf{z}_1,\mathbf{z}_2,...,\mathbf{z}_S,\mathbf{r}_{\mathbf{y}_1}, \mathbf{r}_{\mathbf{y}_2},...,\mathbf{r}_{\mathbf{y}_{t-1}})
\end{equation}
where $f_d(\cdot)$ refers to the decoder.
As a result, to generate a complete report, the above process is repeated until the generation is finished.

\section{Experiment Settings}

\subsection{Datasets}
\label{subsec:datasets}
\begin{table}[t]
\footnotesize
\centering
\setlength{\tabcolsep}{1mm}{\begin{tabular}{@{}l|rrr|rrr@{}}
\toprule
\multirow{2}{*}{\textsc{\textbf{Dataset}}} & \multicolumn{3}{c|}{\textsc{\textbf{IU X-Ray}}}                                                                     & \multicolumn{3}{c}{\textsc{\textbf{MIMIC-CXR}}}                                                                    \\ \cmidrule(l){2-7} 
                                  & \multicolumn{1}{c}{\textsc{Train}} & \multicolumn{1}{c}{\textsc{Val}} & \multicolumn{1}{c|}{\textsc{Test}} & \multicolumn{1}{c}{\textsc{Train}} & \multicolumn{1}{c}{\textsc{Val}} & \multicolumn{1}{c}{\textsc{Test}} \\ \midrule
\textsc{Image \#}                 & 5.2K                              & 0.7K                              & 1.5K                                & 369.0K                            & 3.0K                            & 5.2K                             \\
\textsc{Report \#}                & 2.8K                              & 0.4K                              & 0.8K                                & 222.8K                            & 1.8K                            & 3.3K                             \\
\textsc{Patient \#}               & 2.8K                              & 0.4K                              & 0.8K                                & 64.6K                             & 0.5K                              & 0.3K                               \\
\textsc{Avg. Len.}                & 37.6                              & 36.8                            & 33.6                              & 53.0                              & 53.1                            & 66.4                             \\ \bottomrule
\end{tabular}}
\vskip -0.25em
\caption{The statistics of the two benchmark datasets w.r.t. their training, validation and test sets, including the numbers of images, reports and patients, and the averaged word-based length (\textsc{Avg. Len.}) of reports.}
\label{table:datasets}
\vskip -1em
\end{table}
\begin{table*}[t]
\centering
\setlength{\tabcolsep}{1.33mm}{\begin{tabular}{@{}l|l|ccccccc|ccc@{}}
\toprule
\multirow{2}{*}{\textsc{\textbf{Data}}}                                             & \multicolumn{1}{c|}{\multirow{2}{*}{\textsc{\textbf{Model}}}} & \multicolumn{7}{c|}{\textsc{\textbf{NLG Metrics}}}                                                              & \multicolumn{3}{c}{\textsc{\textbf{CE Metrics}}} \\
                                                                                       & \multicolumn{1}{c|}{}                                         & \textsc{BL-1}  & \textsc{BL-2}  & \textsc{BL-3}  & \textsc{BL-4}  & \textsc{MTR}   & \textsc{RG-L}  & \textsc{Avg. $\Delta$} & \textsc{P}        & \textsc{R}       & \textsc{F1}      \\ \midrule
\multirow{3}{*}{\begin{tabular}[c]{@{}l@{}}\textsc{IU}\\ \textsc{X-Ray}\end{tabular}}  & \textsc{Base}                                              & 0.396          & 0.254          & 0.179          & 0.135          & 0.164          & 0.342          & -                      & -                 & -                & -                \\
                                                                                       & \textsc{~+mem}                                               & 0.443          & 0.270          & 0.191          & 0.144          & 0.172          & 0.351          & \ \ 6.6\%              & -                 & -                & -                \\
                                                                                       & \textsc{~+cmn}                                                 & \textbf{0.475} & \textbf{0.309} & \textbf{0.222} & \textbf{0.170} & \textbf{0.191} & \textbf{0.375} & \textbf{19.6\%}        & -                 & -                & -                \\ \midrule
\multirow{3}{*}{\begin{tabular}[c]{@{}l@{}}\textsc{MIMIC}\\ \textsc{-CXR}\end{tabular}} & \textsc{Base}                                              & 0.314          & 0.192          & 0.127          & 0.090          & 0.125          & 0.265          & -                      & 0.331             & 0.224            & 0.228            \\
                                                                                       & \textsc{~+mem}                                               & 0.340          & 0.209          & 0.140          & 0.100          & 0.135          & 0.273          & \ \ 8.2\%              & 0.322             & 0.255            & 0.261            \\
                                                                                       & \textsc{~+cmn}                                                 & \textbf{0.353} & \textbf{0.218} & \textbf{0.148} & \textbf{0.106} & \textbf{0.142} & \textbf{0.278} & \textbf{13.1\%}        & \textbf{0.334}    & \textbf{0.275}   & \textbf{0.278}   \\ \bottomrule
\end{tabular}}
\vskip -0.3em
\caption{
NLG and CE evaluations of different models on the test sets of \textsc{IU X-Ray} and \textsc{MIMIC-CXR} datasets.
BL-n denotes BLEU score using up to 4-grams; MTR and RG-L denote METEOR and ROUGE-L, respectively.
The average improvement over all NLG metrics compared to \textsc{Base} is also presented in the ``\textsc{Avg. $\Delta$}'' column.}
\label{table:baseline}
\vskip -1em
\end{table*}
We employ two conventional benchmark datasets in our experiments, i.e., \textsc{IU X-Ray} \cite{iuxray}\footnote{\url{https://openi.nlm.nih.gov/}} from Indiana University and \textsc{MIMIC-CXR} \cite{mimic}\footnote{\url{https://physionet.org/content/mimic-cxr/2.0.0/}} from the Beth Israel Deaconess Medical Center.
The former is a relatively small dataset with 7,470 chest X-ray images and 3,955 corresponding reports; the latter is the largest public radiography dataset with 473,057 chest X-ray images and 206,563 reports.

Following the experiment settings from previous studies \cite{hrgr,cmas,r2gen}, we only generate the findings section and exclude the samples without the findings section for both datasets.
For \textsc{IU X-Ray}, we use the same split (i.e., 70\%/10\%/20\% for train/validation/test set) as that stated in \newcite{hrgr} and for \textsc{MIMIC-CXR} we adopt its official split.
Table \ref{table:datasets} show the statistics of all datasets in terms of the numbers of images, reports, patients and the average length of reports with respect to train/validation/test set.

\subsection{Baseline and Evaluation Metrics}
\label{subsec:baselines-and-metrics}
To examine our proposed model, we use the following ones as the main baselines in our experiments:
\begin{itemize}
    \item \textbf{\textsc{Base}}: this is the backbone encoder-decoder used in our full model, i.e., a three-layer Transformer model with 8 heads and 512 hidden units without other extensions.
    \item \textbf{\textsc{Base+mem}}: this is the Transformer model with the same architecture of \textbf{\textsc{Base}} where two memory networks are separately applied to image and text, respectively. This baseline aims to provide a reference to the cross-modal memory.
\end{itemize}
To further demonstrate the effectiveness of our model, we compare it with previous studies, including conventional image captioning models, e.g., \textbf{\textsc{ST}} \cite{showandtell}, \textbf{\textsc{Att2in}} \cite{scst}, \textbf{\textsc{AdaAtt}} \cite{adaatt},  \textbf{\textsc{Topdown}} \cite{updown}, and the ones proposed for the medical domain, e.g., \textbf{\textsc{CoAtt}} \cite{coatt}, \textbf{\textsc{Hrgr}} \cite{hrgr}, \textbf{\textsc{Cmas-Rl}} \cite{cmas} and \textbf{\textsc{R2Gen}} \cite{r2gen}.

Following \newcite{r2gen}, we evaluate the above models by two types of metrics, conventional natural language generation (NLG) metrics and clinical efficacy (CE) metrics\footnote{Note that CE metrics only apply to \textsc{MIMIC-CXR} because the labeling schema of CheXpert is designed for \textsc{MIMIC-CXR}, which is different from that of \textsc{IU X-Ray}.}.
The NLG metrics\footnote{\url{https://github.com/tylin/coco-caption}} include BLEU \cite{bleu}, METEOR \cite{meteor} and ROUGE-L \cite{rouge}.
For CE metrics, the CheXpert \cite{chexpert}\footnote{\url{https://github.com/MIT-LCP/mimic-cxr/tree/master/txt/chexpert}} is applied to label the generated reports and compare the results with ground truths in 14 different categories related to thoracic diseases and support devices.
We use precision, recall and F1 to evaluate model performance for CE metrics.

\subsection{Implementation Details}
\label{subsec:implementation-details}
To ensure consistency with the experiment settings of previous work \cite{hrgr,r2gen}, we use two images of a patient as input for report generation on \textsc{IU X-Ray} and one image for \textsc{MIMIC-CXR}.
For visual extractor, we adopt the ResNet101 \cite{resnet} pretrained on ImageNet \cite{imagenet} to extract patch features with 512 dimensions for each feature.
For the encoder-decoder backbone, we use a Transformer structure with 3 layers and 8 attention heads, 512 dimensions for hidden states and initialize it randomly.
For the memory matrix in CMN, its dimension and the number of memory vectors $\mathcal{N}$ are set to 512 and 2048, respectively, and also randomly initialized.
For memory querying and responding, thread number and the $\mathcal{K}$ are set to 8 and 32, respectively.
We train our model under cross entropy loss with Adam optimizer \cite{adam}.
The learning rates of the visual extractor and other parameters are set to $5\times 10^{-5}$ and $1\times10^{-4}$, respectively, and we decay them by a $0.8$ rate per epoch for all datasets.
For the report generation process, we set the beam size to 3 to balance the effectiveness and efficiency of all models.
Note that the optimal hyper-parameters mentioned above are obtained by evaluating the models on the validation sets from the two datasets.

\section{Results and Analyses}
\label{sec:results-and-analysis}
\subsection{Effect of Cross-Modal Memory}
\begin{table*}[t]
\centering
\begin{tabular}{@{}l|l|cccccc|ccc@{}}
\toprule
\multirow{2}{*}{\textsc{\textbf{Data}}}                                             & \multicolumn{1}{c|}{\multirow{2}{*}{\textsc{\textbf{Model}}}} & \multicolumn{6}{c|}{\textsc{\textbf{NLG Metrics}}}                                     & \multicolumn{3}{c}{\textsc{\textbf{CE Metrics}}} \\
                                                                                       & \multicolumn{1}{c|}{}                                         & \textsc{BL-1}  & \textsc{BL-2}  & \textsc{BL-3}  & \textsc{BL-4}  & \textsc{MTR}   & \textsc{RG-L}  & \textsc{P}     & \textsc{R}    & \textsc{F1}      \\ \midrule
\multirow{7}{*}{\begin{tabular}[c]{@{}l@{}}\textsc{IU}\\ \textsc{X-Ray}\end{tabular}}  & \textsc{ST}$^{\ddag}$                                      & 0.216          & 0.124          & 0.087          & 0.066          & -              & 0.306          & -                 & -                & -                \\
                                                                                       & \textsc{Att2in}$^{\ddag}$                                  & 0.224          & 0.129          & 0.089          & 0.068          & -              & 0.308          & -                 & -                & -                \\
                                                                                       & \textsc{AdaAtt}$^{\ddag}$                                  & 0.220          & 0.127          & 0.089          & 0.068          & -              & 0.308          & -                 & -                & -                \\ \cmidrule(l){2-11} 
                                                                                       & \textsc{CoAtt}$^{\ddag}$                                   & 0.455          & 0.288          & 0.205          & 0.154          & -              & 0.369          & -                 & -                & -                \\
                                                                                       & \textsc{Hrgr}$^{\ddag}$                                    & 0.438          & 0.298          & 0.208          & 0.151          & -              & 0.322          & -                 & -                & -                \\
                                                                                       & \textsc{Cmas-RL}$^{\ddag}$                                 & 0.464          & 0.301          & 0.210          & 0.154          & -              & 0.362          & -                 & -                & -                \\
                                                                                        & \textsc{R2Gen}$^{\ddag}$                                  & 0.470          & 0.304          & 0.219          & 0.165          & 0.187              & 0.371          & -                 & -                & -                \\
                                                                  \cmidrule(l){2-11} 
                                                                                       & \textsc{Ours (cmn)}                                                 & \textbf{0.475} & \textbf{0.309} & \textbf{0.222} & \textbf{0.170} & \textbf{0.191}          & \textbf{0.375} & -                 & -                & -                \\ \midrule
\multirow{5}{*}{\begin{tabular}[c]{@{}l@{}}\textsc{MIMIC}\\ \textsc{-CXR}\end{tabular}} & \textsc{ST}$^{\Diamond}$                                        & 0.299          & 0.184          & 0.121          & 0.084          & 0.124          & 0.263          & 0.249             & 0.203            & 0.204            \\
                                                                                       & \textsc{Att2in}$^{\Diamond}$                                    & 0.325          & 0.203          & 0.136          & 0.096          & 0.134          & 0.276          & 0.322             & 0.239            & 0.249            \\
                                                                                       & \textsc{AdaAtt}$^{\Diamond}$                                    & 0.299          & 0.185          & 0.124          & 0.088          & 0.118          & 0.266          & 0.268             & 0.186            & 0.181            \\
                                                                                       & \textsc{Topdown}$^{\Diamond}$                                   & 0.317          & 0.195          & 0.130          & 0.092          & 0.128          & 0.267          & 0.320             & 0.231            & 0.238            \\ 
                                                                                       & \textsc{R2Gen}$^{\ddag}$                                   & 0.353          & 0.218          & 0.145          & 0.103          & 0.142          & 0.277          & 0.333             & 0.273            & 0.276                \\                            \cmidrule(l){2-11} 
                                                                                       & \textsc{Ours (cmn)}                                                 & \textbf{0.353} & \textbf{0.218} & \textbf{0.148} & \textbf{0.106} & \textbf{0.142} & \textbf{0.278} & \textbf{0.334}    & \textbf{0.275}   & \textbf{0.278}   \\ \bottomrule
\end{tabular}
\vskip -0.3em
\caption{Comparisons of our proposed model with previous studies on the test sets of \textsc{IU X-Ray} and \textsc{MIMIC-CXR} with respect to NLG and CE metrics.
 ${\ddag}$ refers to that the result is directed cited from the original paper and ${\Diamond}$ represents our replicated results by their released codes.}
\label{table:sota}
\vskip -1em
\end{table*}
The main experimental results on the two aforementioned datasets are shown in Table \ref{table:baseline}, where \textsc{Base+cmn} represents our model (same below).
There are several observations drawn from different aspects.
First, both \textsc{Base+mem} and \textsc{Base+cmn} outperform the vanilla Transformer (\textsc{Base}) on both datasets with respect to NLG metrics, which confirms the validity of incorporating memory to introduce more knowledge into the Transformer backbone.
Such knowledge may come from 
the hidden structures and regularity patterns shared among radiology images and their reports, so that the memory modules are able to explicitly and reasonably model them to promote the recognition of diseases (symptoms) and the generation of reports.
Second, the comparison between \textsc{Base+cmn} and two baselines on different metrics confirms the effectiveness of our proposed model with significant improvement.
Particularly, \textsc{Base+cmn} outperforms \textsc{Base+mem} by a large margin, which indicates the usefulness of CMN in learning cross-modal features with a shared structure rather than separate ones.
Third, when comparing between datasets, the performance gains from \textsc{Base+cmn} over two baselines (i.e., \textsc{Base} and \textsc{Base+mem}) on \textsc{MIMIC-CXR} are larger than that of \textsc{IU X-Ray}.
This observation owes to the fact that \textsc{MIMIC-CXR} is relatively larger, which helps the learning of the alignment between images and texts so that CMN helps more on report generation on \textsc{MIMIC-CXR}.
Third, when compared between datasets, the performace gain from \textsc{Base+cmn} over two baselines (i.e., \textsc{Base} and \textsc{Base+mem}) on \textsc{IU X-Ray} are larger than that of \textsc{MIMIC-CXR}.
This observation owes to the fact that IU X-Ray is relatively small and has less complicated visual-textual mappings,
thus easier for generation by CMN.
Moreover, this size effect also helps that our model shows the same trend on the CE metrics on \textsc{MIMIC-CXR} as that for NLG metrics, where it outperforms all its baselines in terms of precision, recall and F1.

\subsection{Comparison with Previous Studies}
To further demonstrate the effectiveness, we further compare our model with existing models on the same datasets, with their results reported in Table \ref{table:sota} on both NLG and CE metrics.
We have following observations.
First, cross-modal memory shows its effectiveness in this task, where our model outperforms \textsc{CoAtt}, although both of them improve the report generation by the alignment of visual and textual features.
The reason behind might be that our model is able to use a shared memory matrix as the medium to softly align the visual and textual features instead of direct alignment using the co-attention mechanism,
thus unifies cross-modal features within same representation space and facilitate the alignment process.
Second, our model confirms its superiority of simplicity when comparing with those complicated models.
For example, \textsc{Hrgr} uses manually extracted templates and \textsc{Cmas-RL} utilizes reinforcement learning with a careful design of adaptive rewards and our model achieves better results with a rather simpler method.
Third, applying memory to both the encoding and decoding can further improve the generation ability of Transformer when compared with \textsc{R2Gen} which only uses memory in decoding.
This observation complies with our intuition that the cross-modal operation tightens the encoding and decoding so that our model generates higher quality reports.
Fourth, note that although there are other models (i.e., \textsc{CoAtt} and \textsc{Hrgr})
with exploiting extra information (such as private datasets for visual extractor pre-training), our model still achieves the state-of-the-art performance without requiring such information.
It reveals that in this task, the hidden structures among the images and texts and a good solution of exploiting them are more essential in promoting the report generation performance.

\begin{figure}[t]
\centering
\includegraphics[width=0.48\textwidth, trim=0 25 0 10]{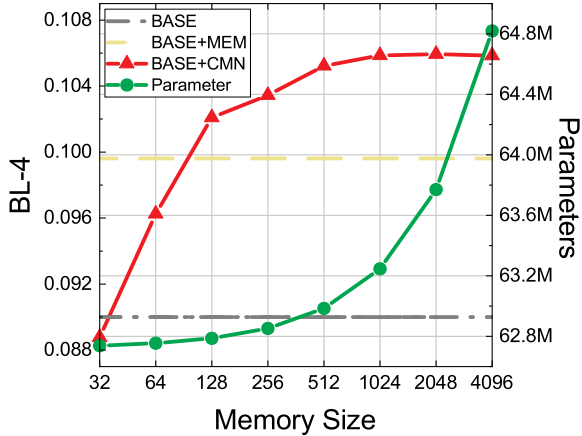}
\caption{The BLEU-4 score and the number of parameters from \textsc{Base+cmn} against the memory size (i.e., number of memory vectors) when the model is trained and tested on \textsc{MIMIC-CXR} dataset.}
\label{fig:memory-size}
\vskip -1em
\end{figure}

\subsection{Analysis}
\paragraph{Memory Size}
To analyze the impacts of memory size, we train our model with different numbers of memory vectors, i.e.,
$\mathcal{N}$ ranges from 32 to 4096,
with the results on \textsc{MIMIC-CXR} shown in Figure \ref{fig:memory-size}.
It is observed that,
first, enlarging memory by the number of vectors results in better overall performance when the entire memory matrix is relatively small ($\mathcal{N} \leq 1024$),
which can be explained by
that, within a certain memory capacity,
larger memory size helps store more cross-modal information;
second, when the memory matrix is larger than a threshold, increasing memory vectors is not able to continue promising a better outcome.
An explanation to this observation may be that, when the matrix is getting to large, the memory vectors can not be fully updated so they do not help the generation process other than being played as noise.
More interestingly, it is noted that
even if we use a rather large memory size (i.e., $\mathcal{N} = 4096$), only 3.34\% extra parameters are added to the model compared to \textsc{Base},
which justifies that introducing memory to report generation process through our model can be done with small price.

\begin{figure}[t]
\centering
\includegraphics[width=0.48\textwidth, trim=0 25 0 10]{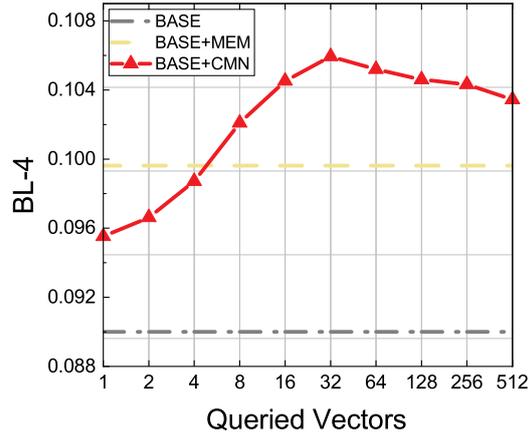}
\caption{The BLEU-4 score from \textsc{Base+cmn} when tested on the \textsc{MIMIC-CXR} test set against different numbers of queried memory vectors.}
\label{fig:addressed-keys}
\vskip -1em
\end{figure}

\paragraph{Number of Queried Memory Vectors}
To analyze how querying impacts report generation,
we try CMN with different numbers of queried vectors, i.e., $\mathcal{K}$ ranges from 1 to 512,
and show the results in Figure \ref{fig:addressed-keys}.
It is found that
the number of queried vectors should be neither too small nor too big,
where enlarging $\mathcal{K}$ leads to better results when $\mathcal{K} \leq 32$ and after this threshold the performance starts to drop.
The reason behind might be the overfitting of memory updating since the memory matrix is sparsely updated in each iteration when $\mathcal{K}$ is small, i.e., it is hard to be overfit under this scenario, while more queried vectors should cause intensive updating on the matrix and some of the essential vectors are over-updated accordingly.
As a result, it is interesting to find the optimal number (i.e., 32) of queried vectors and this is a useful guidance to further improve report generation with controlling the querying process.

\begin{figure*}[t]
\centering
\includegraphics[width=1\textwidth, trim=0 20 0 10]{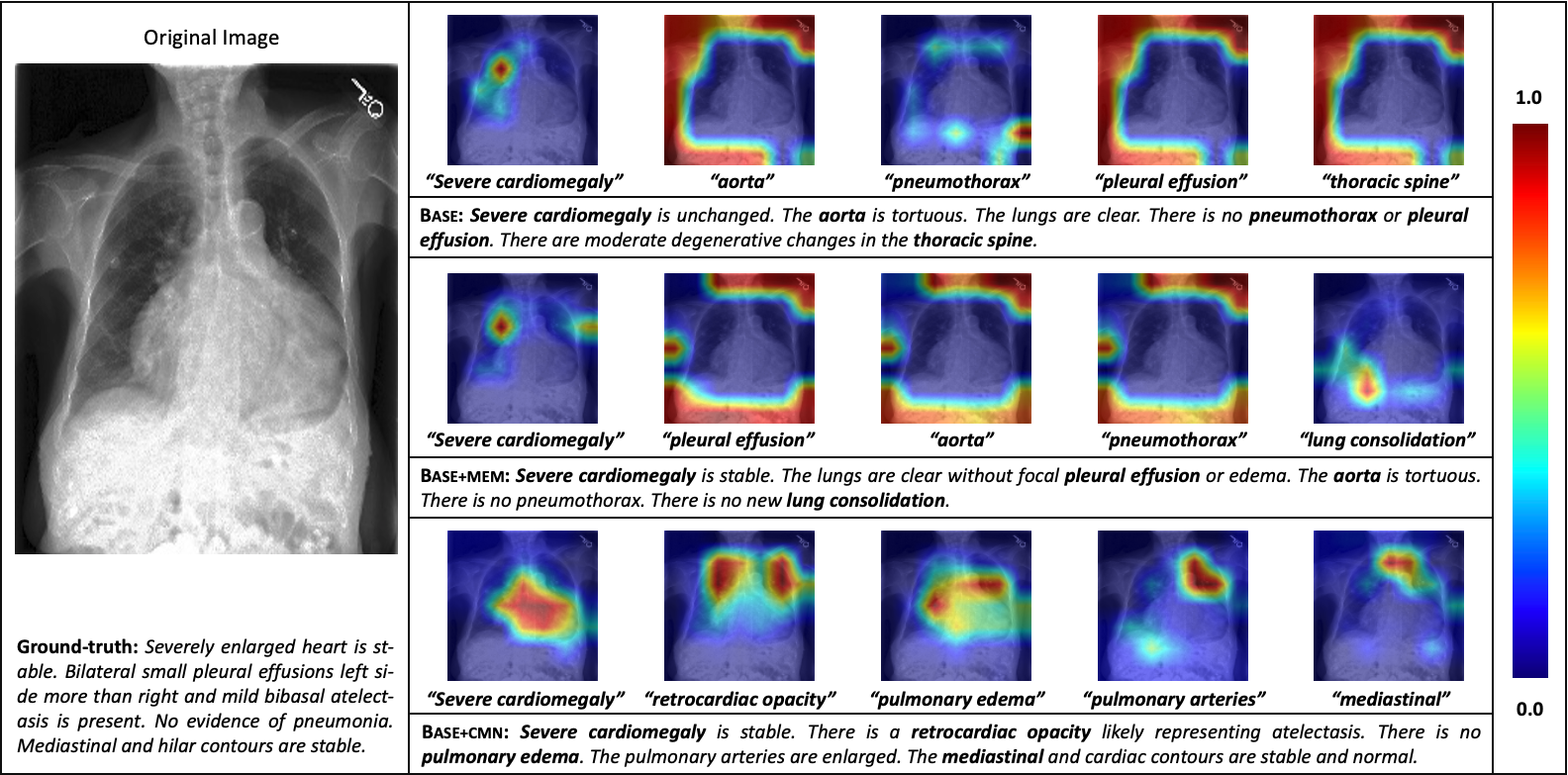}
\caption{Visualizations of image-text mappings between particular regions (indicated by colored weights) of a chest X-ray image and words/phrases from its reports generated by \textsc{Base}, \textsc{Base+mem} and \textsc{Base+cmn}, respectively.
The color spectrum indicates the value of weight from low to high in the range of [0, 1].
}
\label{fig:attention}
\vskip -1.15em
\end{figure*}
\begin{figure}[t]
\centering
\includegraphics[width=0.48\textwidth, trim=0 20 0 5]{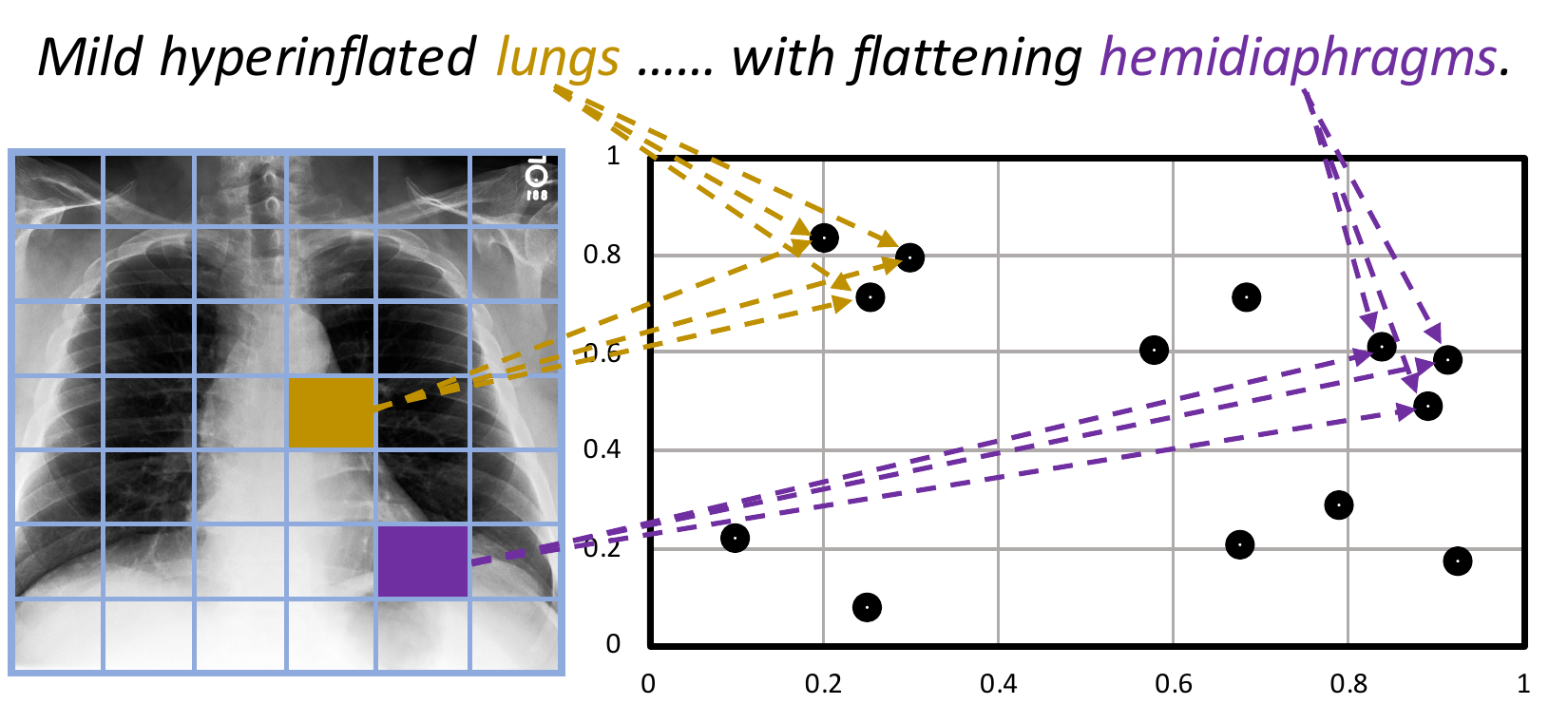}
\caption{T-SNE visualization of memory vectors with an example input image and its partial generated report from \textsc{MIMIC-CXR} test set. The queried vectors for visual and textual features are indicated by arrows.}
\label{fig:tsne}
\vskip -1em
\end{figure}

\paragraph{Case Study}
To further qualitatively investigate how our model learns from the alignments between the visual and textual information, we perform a case study on the generated reports from different models regarding to an input chest X-ray image chosen from \textsc{MIMIC-CXR}.
Figure \ref{fig:attention} shows the image with ground-truth report, and different reports with selected mappings from visual (some part of the image) and textual features (some words and phrases),\footnote{The representations of the textual features are extracted from the first layer of the decoder.} where the mapped areas on the image are highlighted with different colors.
In general,
\textsc{Base+cmn} is able to generate more accurate descriptions (in terms of better visual-textual mapping) in the report while other baselines are inferior in doing so.
For instance, normal medical conditions and abnormalities presented in the chest X-ray image are covered by the generated report from \textsc{Base+cmn} (e.g., ``\textit{severe cardiomegaly}'', ``\textit{pulmonary edema}'' and ``\textit{pulmonary arteries}'') and the related regions on the image are precisely located regarding to the texts,
while the areas highlighted on the image from other models are inaccurate.

To further illustrate how the alignment works between visual and textual features, we perform a t-SNE visualization on the memory vectors linking to an image and its generated report from the \textsc{MIMIC-CXR} test set.
It is observed that the word ``\textit{lung}'' in the report and the visual feature for the region of lung on the image query similar memory vectors from CMN,
where similar observation is also drawn for ``hemidiaphragms'' and its corresponding regions on the image.
This case confirms that memory vector is effective intermediate medium to interact between image and text features.

\section{Related Work}
\label{sec:relatedwork}
In general, the most popular related task to ours is image captioning, a cross-modal task involving natural language processing and computer vision, which aims to describe images in sentences \cite{showandtell,xu2015show,updown,wang2019hierarchical,m2}.
Among these studies, the most related study from \citet{m2} also proposed to leverage memory matrices to learn a priori knowledge for visual features using memory networks \cite{memory,end2end,zeng2018topic,relational,nie2020improving,diao2020keyphrase,tian2020improving,tian-etal-2021-enhancing,chen-etal-2021-relation}, but such operation is only performed during the encoding process.
Different from this work, the memory in our model is designed to align the visual and textual features, and the memory operations (i.e., querying and responding) are performed in both the encoding and decoding process.

Recently, many advanced NLP techniques (e.g., pre-trained language models) have been applied to tasks in the medical domain \cite{pampari2018emrqa,zhang2018learning,wang2018coding,alsentzer2019publicly,tian2019chimed,tian-etal-2020-improving-biomedical,wang-etal-2020-studying,lee2020biobert,song2020summarizing}.
Being one of the applications and extensions of image captioning to the medical domain, radiology report generation aims to depicting radiology images with professional reports.
Existing methods were designed and proposed to better align images and texts or to exploit highly-patternized features of texts.
For the former studies, \newcite{coatt} proposed a co-attention mechanism to simultaneously explore visual and semantic information with a multi-task learning framework.
For the latter studies, \newcite{hrgr} introduced a template database to incorporate patternized information and \newcite{r2gen} improved the performance of radiology report generation by applying a memory-driven Transformer to model patternized information.
Compared to these studies, our model offers an effective yet simple alternative to generating radiology reports,
where a soft intermediate layer is provided to facilitate the mappings between visual and textual features, so that more accurate descriptions are produced for generation.

\section{Conclusion}
In this paper, we propose to generate radiology reports with cross-modal memory networks,
where a memory matrix is employed to record the alignment and interaction between images and texts,
with memory querying and responding performed to obtain the shared information across modalities.
Experimental results on two benchmark datasets demonstrate the effectiveness of our model, which achieves the state-of-the-art performance.
Further analyses investigate the effects of hyper-parameters in our model and show that our model is able to better align information from images and texts, so as to generate more accurate reports, especially with the fact that enlarging the memory matrix does not significantly affect the entire model size.

\section*{Acknowledgments}
This work is supported by Chinese Key-Area Research and Development Program of Guangdong Province (2020B0101350001) and NSFC under the project ``The Essential Algorithms and Technologies for Standardized Analytics of Clinical Texts'' (12026610).

\bibliographystyle{acl_natbib}
\bibliography{acl2021}

\end{document}